\documentstyle[12pt]{article}
\newcommand{\anf}[1]{`{#1}'}

\newcommand{\mylist}[1]{\begin{list}{$\bullet$}{\leftmargin4mm 
\labelwidth2mm \labelsep1mm \itemsep0mm \parsep1mm} #1 \end{list}}
\newcommand{\ak}[1]{#1}
\newcommand{\mybib}[1]{\noindent\vspace{-1mm}\hfill\begin{minipage}[t]{7.6cm}\hspace{-4mm}#1\end{minipage}\\}

\begin{document}
\bibliographystyle{g_xalpha}

\renewcommand{\thefootnote}{\fnsymbol{footnote}}

\twocolumn[\begin{center}
\vspace*{-12mm}
{\scriptsize Appears in the {\em Proceedings of the 15th International Conference
on Computational Linguistics\\ (COLING 94)}, Kyoto, Japan, August 1994, pp.\ 356
-- 362}\\
\vspace{5mm}
{\large\bf DEFAULT HANDLING IN INCREMENTAL GENERATION}\\
\vspace{6mm} Karin Harbusch$\dagger$, Gen--ichiro Kikui$\ddagger$,  Anne
Kilger$\dagger$\\ 
\vspace{4mm}
$\ddagger$ ATR\footnotemark, 
Fax: (+81 7749) 5 1308, E-mail: kikui@itl.atr.co.jp\\
$\dagger$ DFKI,
Fax: (+49 681) 302 5341, E-mail: harbusch$\vert$kilger@dfki.uni-sb.de
\end{center}
\vspace{3mm}
]

{\bf Abstract}
{\small  Natural language generation must work with insufficient input.
Underspecifications can be caused by shortcomings of the component  providing the
input or by the preliminary state of incrementally given input. The paper aims
to escape from such dead--end situations by making assumptions. We discuss global
aspects of default handling. Two problem classes for defaults in the
incremental syntactic generator VM--GEN are presented to substantiate
our discussion. }

\footnotetext{The author is currently at NTT Network Information Systems
Laboratories (kikui@nttnly.ntt.jp).}

\renewcommand{\thefootnote}{\arabic{footnote}}
\setcounter{footnote}{0}


\vspace{-3mm}
\section*{\protect\large\bf 1 MOTIVATION}
\label{sec-1}
\vspace{-2mm}

{\it Natural Language Generation}, i.e., the process of building
an adequate utterance for\linebreak some given content, is by nature a decision--making
problem (Appelt, 1985). Internal decisions are made on the basis of the
specified input.  Unfortunately, input information can be insufficient in two
respects:  

$\bullet$ If the input structure for generation is provided by another
AI--system, {\em global problems in producing sufficient input information} for
the generator may occur, e.g., because of\linebreak translation mismatches in machine
translation (Kameyama, 1991). In this case, the generator 
either has to use a default or formulate a request for clarification
in order to be able to continue its processing, i.e., to produce an
utterance. During simultaneous interpretation requests are rather unusual.
Here defaults allow for a standalone handling of the problem. For example,
problems during\linebreak speech recognition of automatic interpretation can lead to
results like ``the (man/men) will come to the hotel tomorrow''. If the system
is not able to give a preference for one of the alternatives, e.g., by
evaluating context information, the generator has to choose a probable
number value on its own to complete verbalization.

$\bullet$ Furthermore, for incremental generation,\linebreak the input
information is produced 
and handed over step by	step, so that it can be {\em temporarily incomplete} ---
although as a whole it may become sufficient. This behaviour of a
generator is motivated by psycholinguistic observations which show that	people
start speaking before all necessary linguistic material	has been chosen (e.g.,
articulating a noun phrase before the dominating verb is selected). 
As a consequence of underspecification, incremental generation is essentially
based on working with defaults.	Elements are uttered before the processing or
input  consumption has been finished. (Kitano, 1990) gives an example for 
defaults in the context of simultaneous interpretation: In Japanese, negation
is specified at the end of the sentence while in English, it has to be
specified in front of the finite verb. Therefore, during Japanese--English
translation, where analysis, transfer, and generation are performed in a
parallel and incremental way, the system has to commit, e.g., positive
value before knowing the actual polarity\footnote{Alternatively, the
system could use the dialogue context to infer a negation value +/-.}.

Generally speaking, {\it default handling} specifies how processing, i.e.,
further decision--mak\-ing, can continue without  sufficient input information. So,
one can compare default handling with {\it advice to the system}. For
reasons\linebreak of uncertainty of assumptions, incremental\linebreak systems
with this facility  
must be able to repair the default decision when the assumption turns out to be
wrong by information given later. Catching on to the above example, there
can be a negation specifier given at the end of the Japanese input sentence
which cannot be simply integrated into the output sentence because the finite
verb has already been uttered. In this case, the output has to be repaired,
e.g., by repeating parts of the utterance: ``I will be able to meet you \ldots
oops \ldots I won't be able to meet you at the hotel this evening.''

In the following sections, we argue for the appropriateness of {\it
pro\-cess\-ing--con\-form\-ing} default handling. Basically, the
pro\-cess\-ing--con\-for\-ming mode  makes the overall system {\it
homogeneous} because the combination of\linebreak
de\-fault--caused processing and input--licensed processing requires no specific
description.\linebreak The homogeneity becomes especially helpful in the  case where
the input verifies the default assumption  rendering unnecessary any
recomputation. For the opposite case where the default must be withdrawn we
have to mark all defaults. Even more homogeneity is introduced to an incremental
system if the default descriptions are given {\em in terms of input specifications}.
This representation allows for easy checking the coincidence between a chosen
default and input given later.

The content of this paper can be summarized as follows. Section~2
provides a general description for defaults in generation emphasizing the
specific requirements in an incremental system. After identifying
the conditions under which defaults are triggered (section~2.1), the
application of a default (section~2.2) and 
the definition of its description (section~2.3) is outlined. The
crucial case of removing defaults not coinciding with newly arriving input in
an incremental system is discussed in section~2.4.

In section~3, this mechanism is applied to the incremental sentence
generator VM--GEN. In the beginning of the section, the basic design of the
system is outlined. Later on, default handling is included and exemplified
for two general cases. 

In the final section we summarize the main results of the paper. Furthermore,
we discuss how default handling can be adapted to {\em multilingual
generation}, as required by the speech--to--speech translation system VERBMOBIL
(Block et al., 1992).

\vspace{-3mm}
\section*{\protect\large\bf 2 GENERAL DISCUSSION OF DEFAULTS}
\label{sec-2}
\vspace{-2mm}

In the literature of non--incremental generation, the need for defaults
is hardly ever taken into account. The common point of view restricts the 
input to be sufficient for generation (see, e.g., the
{\em Text Structure} by (Meteer, 1990) for a syntactic generator). In
incremental generation, most authors agree on the necessity of using defaults
(see, e.g., (De~Smedt, 1990; Kitano, 1990; Ward, 1991)). Nevertheless,
they do not in sufficient depth answer the question of how to guide the
processes of default handling and repair within a generator. This problem is
the starting--point for the following considerations.

We assume that generation is a decision--making process with the aim of
producing a plausible utterance based on  given information. As mentioned
in section~1, there are cases where this process stops (caused by
underspecification of the input) before finishing its output.  

We define a module named {\em default handler} which tries to resume the
process by giving advice to it, i.e., by making assumptions about the missing
input specification. With respect to this task it is discussed   
\mylist{
\vspace{-2mm}
\item[1.] in which situations defaults are applied (see
section~2.1),  
\item[2.] how default handling is integrated into a system (see
section~2.2),   
\item[3.] how the knowledge for default handling is
described (see section~2.3), and  
\item[4.] how assumptions are cancelled when they turn out to be inconsistent
with newly arriving input (see section~2.4).  
}
\vspace{-2mm}
In incremental generation, as mentioned in section~1, interleaved input
consumption and output production causes specific default situations. An
incremental processing scheme allows for an increase of efficiency and
flexibility, e.g., by making the analysis and generation processes of a system
for simultaneous interpretation overlap in time. There are two competing goals
of incremental generation for spoken output, that must be taken into account
when estimating the usefulness of defaults:

\mylist{
\vspace{-2mm}
\item[\bf Fluency:] Long hesitations should be avoided  during the production
of an utterance, in order to be {\it acceptable}
 to the hearer\footnote{Humans often
fill such pauses with {\em fillers} like ``er''  or ``what shall I say''.}.

\item[\bf Reliability:] Errors in an utterance may\linebreak cause misunderstanding.
In most cases, errors should be recovered by appropriate
self--cor\-rec\-tions\footnote{Sometimes, correction is unnecessary if (the
speaker believes that) the hearer can infer the intended utterance from
erroneous speech.}. Excessive use of self--cor\-rec\-tions or erroneous
expressions should be\linebreak avoided because they decrease {\it 
intelligibility} of the
utterance.  

}
\vspace{-2mm}
Obviously there is a trade--off between fluency and reliability: 
maximal reliability requires `secure' decisions and therefore leads to
output delay. On the other hand, maximal fluency necessitates the use
of assumptions and repair, respectively.

\vspace{-3mm}
\subsection*{\protect\normalsize\bf 2.1 When to Trigger Default Handling}
\label{sec-2.1}
\vspace{-2mm}

We define as {\em default situation} the situation where a generation system
has not yet finished the utterance but at the same time has consumed all given
input and is not able to continue processing. In non--incremental generation,
this corresponds to the fact that  the input lacks necessary information,
because\linebreak the entire input is assumed to be given at one time \ak{(e.g., the
undecidable number value of the example described in section~1)}. Thus,
default handling should be triggered immediately. 

In incremental generation, however, the system may get a new piece of
information later on that enables it to continue processing \ak{(e.g., the
specification of a negation value + as outlined in the example in section~1).}
Therefore, possible alternatives are either to wait for 
the next input or to trigger default handling. The former violates the fluency
goal, the latter may violate the reliability goal. We propose the explicit use
of {\em time--limits for delay intervals}\footnote{An explicit parameter
expressing the desired degree of fluency influences the time--limits.}. 

Furthermore, the {\em certainty of a default} is described by a value. As soon as
a  default situation is identified, the certainty of the default is checked to
see whether it exceeds a predefined threshold that determines the degree of
fluency/reliability\footnote{\ak{The basis for assigning certainty values to defaults
should be a corpus study that can be used to find statistical evidence for
various features with alternative values (like number, voice, \ldots, see,
e.g., (Bock and Warren, 1985)).}}. 

Each application of a default decreases the global certainty of the 
system's state. Consequently, there should be a {\em limit for the maximal number
of defaults} applicable to the same sentence.

\vspace{-3mm}
\subsection*{\protect\normalsize\bf 2.2 How to Integrate Default Handling}
\label{sec-2.2}
\vspace{-2mm}

\noindent
Basically, there are two strategies to integrate default handling into ongoing
processing.

Defaults may be handled in a way that differs from the `normal'
processing of the system, e.g., as short--cuts. One advantage can be an
efficient handling of defaults. Furthermore, the designer of the default
component is completely free in deciding about the realization of defaults in the
system. A disadvantage is the difficulty of providing consistency between
de\-fault--caused and input--licensed processing.

Alternatively, the ongoing processing can deal with the default
values in an ordinary manner (pro\-cess\-ing--con\-form\-ing default handling).
This may be less efficient but guarantees consistency during
processing, especially in case of a replacement by an input--licensed
value. For incremental generation, the system has to provide
repair facilities in any case. So, they can also be used for {\it non--mon\-oton\-ic}
modifications of de\-fault--caused results. We take this option in order to make
the overall system {\em   homogeneous}.

\vspace{-3mm}
\subsection*{\protect\normalsize\bf 2.3 How to Describe Defaults}
\label{sec-2.3}
\vspace{-2mm}

The knowledge source that is used for  default handling  should provide the
most plausible actions for a default situation.  We represent the knowledge
as a set of heuristic rules called {\em  default descriptions}. A default
description defines a set of operations that  should be carried out in a
certain situation where the generation process can not be continued. A
default description has the following form: 
\vspace{-2mm}
\begin{center}
\begin{math}
\left[
\begin{array}{l}default\\ preconditions\end{array}
 \Rightarrow
\begin{array}{l}default\\ body\end{array} ; 
\begin{array}{l}certainty\\ value\end{array}
\right]
\end{math}
\end{center}
\vspace{-2mm}
The set of {\em default preconditions} defines tests that are applied to the
given situation in order to find out whether the corresponding default body
can be activated. They include tests for the existence of  particular
information, tests for the structure under creation and tests for the state
of processing. 
 
The {\em default body} describes how to continue processing with defaults in
an adequate way. For incremental systems, we propose to express the
body as a specification of input increments. An important prerequisite is that
the size of increments is defined flexibly enough to cope with varying amounts of
information. Obviously, an important advantage of this  approach is homogeneity
of the overall system. Especially, the homogeneous representation of
de\-fault--caused and input--licensed structures is the easiest and most direct way
to test coincidences or contradictions between de\-fault--spe\-cified and
input--caused values. In section~3, this approach is outlined by
different examples. For non--incremental systems, an operational approach is
preferable since there is no way to consume additional input increments,
presupposing that the input has been considered as a whole before a default
situation occurs\footnote{The difference between incremental and non--incremental
generation becomes smaller, if we assume that defaults in  a non--incremental
system can be triggered after the system has only considered parts of its input
information. In this case, the input considered after default handling becomes
comparable to later increments.}.

If several default preconditions are applicable, the {\em certainty values}
for default descriptions are examined to find which provides the system with the
most plausible action.

The individual default descriptions should take into account the global
constraints for processing stated in the knowledge sources of the system. 
For example, the assumption of nominative case for a German NP complement can
regularly be made only once for the same verb. For reasons of homogeneity, the
default description should at least be compatible with the specifications of
the knowledge used for basic processing. In order to guarantee consistency, 
default descriptions should merely contain what is orthogonal to the basic
knowledge sources.

\vspace{-3mm}
\subsection*{\protect\normalsize\bf 2.4 How to Cancel Defaults}
\label{sec-2.4}
\vspace{-2mm}

The repair of false assumptions is a crucial point of default handling in the
context of  incremental processing \ak{because the default information 
does not remain locally but can cause further decisions of the system}. 
Contrarily, for non--incremental input there will be no value given that can
contradict default values.  

As a first step of repair,  inconsistencies between  input--pro\-vided  and 
de\-fault--caused values are identified by simply matching the values.
Then effects of the respective defaults are
withdrawn introducing the input--provided values into the system.
Generally, a decision during generation influences other decisions all over
the system. Thus the effect of a default body may be propagated through the
entire system (e.g., choosing a construction of main clause with causal
subordinate clause influences the choice of syntactic realizations).

Roughly speaking, withdrawing a default assumption can be realized by
{\it backtracking} to the earlier state of the system where the default had
been introduced or by {\it non--mon\-oton\-ic changes}
to the current state of the system. The disadvantage of backtracking is that
partial results are thrown away which could be reused during further
processing. Non--mon\-oton\-ic changes preserve these results. In this
framework, cancelling defaults requires the system to identify which results
are caused by default handling. {\em Dependency links} between the immediate
result of a default body and results of the influenced decisions allow for
this identification. The disadvantage of non--monotonic changes is the
complexity of computation, 
e.g., supported by a truth maintenance system. When designing an incremental
system, simple backtracking is ruled out because the part of the sentence
uttered cannot be withdrawn after it has been perceived by the addressee of
the message\footnote{If some phrases influenced by defaults have
already been verbalized, the effect of verbalization can be cancelled by 
using repair words like ``oops'' or ``sorry'' when starting the modified
utterance.}.

\medskip
So, we end up with a pro\-cess\-ing--con\-form\-ing default handler for
generation realizing repair by non--mon\-oton\-ic changes. 

\vspace{-3mm}
\section*{\protect\large\bf 3 EXAMPLES OF DEFAULTS IN VM--GEN}
\label{sec-3}
\vspace{-2mm}

The adaptation of our general discussion of default handling to the system
VM--GEN not only provides concrete examples for the reader but also
shows that a homogeneous combination of default handling, \ak{regular
processing, and utterance repair is possible.}

The syntactic generator VM--GEN is a further development of TAG--GEN
(Kilger, 1994) within the framework of VERBMOBIL, a\linebreak speech--to--speech
translation system. Its usefulness for simultaneous interpretation results
from its {\em incremental and parallel style of processing}. VM--GEN is able
to consume input increments of varying size. These increments describe lexical
items or semantic relations between them. Single input increments are handed over
to objects of a distributed parallel system, each of which tries to verbalize
the structure that results from the corresponding input increment. VM--GEN uses
an extension of {\em Tree Adjoining Grammars} (TAGs, cf. (Joshi, 1985)) as its
syntactic representation formalism that is not only adequate for the
description of natural language but also supports incremental generation
(Kilger and Finkler, 1994).

In the following, we introduce examples for default processing triggered
during the German {\em inflection process} in VM--GEN to substantiate the global
statements made in section~2. Inflection uses some syntactic
properties of an element to compute its morphological form. This information
has partly to be specified in the input (e.g., the number for a noun) and is
partly inherited from other elements (e.g., the number for a verb or the case
for a noun). The two reasons for missing information necessitate
different methods of treatment which nevertheless both can uniformly be
integrated into regular processing. 

If information of the first type is missing \ak{(e.g., because of problems during
analysis, see section~1)}, an assumption can be made {\it locally} by
simulating the respective part of the input. The default for missing number 
information in VM--GEN would look as follows:
\vspace{-1mm}  
\begin{center}
\begin{math}
\left[
\begin{array}{l}
{\scriptstyle (cat(OBJ)=N)} \\ {\scriptstyle (number(OBJ)=NIL)}
\end{array}  \Rightarrow 
\begin{array}{r}
{\scriptstyle (ENTITY~OBJ} \\ {\scriptstyle (number~sg))}
\end{array} ;
0.8
\right]\footnotemark
\end{math}\footnotetext{`ENTITY' introduces information about a
lexical  item. For reasons of incrementality, there may be several
ENTITY--packages specified for the same item which are composed to receive the
global information. For certainty values, we use values between 0 and 1,
where 1 means high reliability.}
\end{center}
\vspace{-1mm}
The set of default preconditions is applied to all\footnote{In the actual
implementation we preselect candidates with missing values for reasons of
efficiency.} objects ({\small OBJ}) of VM--GEN in order to test the kind of
underspecification (\anf{number} in the example). 
The default body introduces a new value (sg) by
creating an input increment as a default. The test for coincidence with the
input--licensed value is realized by a comparison in the objects of VM--GEN.
There is a unique association of input increments and objects of VM--GEN (OBJ
is used as identifier) that allows for translating an input modification into
a modification of the state of the respective object. In case of
contradictions the default and all default--caused decisions are 
revised\footnote{For ongoing work on repair in VM--GEN see (Finkler, 1994).}
(see below). 

Making an assumption can be influenced by {\it global} constraints. 
An example, which is well studied in psycholinguistics, is the utterance of a
noun before the verb has been 
chosen. If, e.g., the noun ``Besucher'' (English: ``visitor'') is known to be
the agent of an action, it may be uttered as subject in the first position of
the sentence by default. This treatment presupposes the choice of a `dummy'
verb, which at least subcategorizes a subject and has active
voice\footnote{This kind of expansion is called ``provisional upward
expansion'' by (De~Smedt, 1990).}. The use of a dummy verb and an
underspecified verbal structure the NP is integrated into allows for a simple
global test that rules out the same case value assignment to different NP
complements as it is required for most of the German verbs.
This rule is represented in the grammar as a part of the description of
subcategorization frames for verbs. For reasons of homogeneity we use the
information stored in the syntactic knowledge sources of VM--GEN for
expressing syntactic constraints during default
handling as well.  The advantage of this approach is, that processing is
continued in a consistent way, which eases the introduction of the
input--licensed value. One default for choosing a missing case--value is
specified as follows: 
\vspace{-3mm}
\begin{center} 
\begin{math}
\left[
\begin{array}{l}
{\hspace{-2mm}\scriptstyle (cat(OBJ)=N)} \\
{\hspace{-2mm}\scriptstyle (case(OBJ)=NIL)} \\
{\hspace{-2mm}\scriptstyle (function(OBJ)=agent)} \\
{\hspace{-2mm}\scriptstyle (lemma(head(OBJ))=NIL)} \\
\end{array}
\hspace{-6mm}\Rightarrow
\begin{array}{l}
{\hspace{-2mm}\scriptstyle (ENTITY~OBJ'} \\
 {\scriptstyle (CAT~v)} \\
 {\scriptstyle (VOICE~active))} \\
{\hspace{-2mm}\scriptstyle (RELATION~REL} \\
 {\scriptstyle (HEAD~OBJ')} \\
 {\scriptstyle (MODIFIER~OBJ))}
\end{array}
\hspace{-6mm} ; \hspace{-1mm} 0.8
\right]
\end{math}\footnote{`RELATION' introduces
the specification of a relation between two lexical items which are
identified by the names of their objects.}
\end{center}
\vspace{-2mm}
The default preconditions of the rule characterize a situation where an object
({\small OBJ}) contains no information about the case but identifies the input
category as `N' for noun. Furthermore,
the semantic function of the object is specified as `agent' but no verb
defined yet (lemma(head(OBJ))=NIL) in the head object. That is why, the 
N--object cannot 
inherit a case value and also does not know whether it is allowed to occupy
the front position in the utterance.

Evaluating the default body, the system creates a V--object OBJ'. On the basis
of the input information in (ENTITY OBJ' \ldots) it 
chooses a minimal syntactic structure from the inheritance net of the grammar,
that just desribes a verb category without concrete filler (a dummy verb) plus
a subject complement and active voice for the verbal phrase. Now, the
N--structure is combined with the V--struc\-ture 
of the introduced V--object as during normal processing. Therefore, the
case value can be inherited. Additionally, the first position can be assigned to
the subject which can be uttered now.

\ak{The basic VM--GEN module provides repair strategies
in order to allow for the specification of additions, modifications 
and deletions of input increments, i.e., to model a flexible input interface.
Three features of the system are basically used for repair: First, input
increments are uniquely associated with objects of VM--GEN, so that input
modifications can be translated into modifications of the objects' states.
Second, each modification of an object's state makes it compare new and old
information. In case of a difference, the modified parts are sent to all
concerned objects. Third, the dependency relations that determine the
communication links between objects allow for a hierarchical organization of the
objects, which is the basis for synchronizing repair.

A repair must be triggered in the example described above if, e.g., a verb
with voice passive is actually specified. In this case, the mapping of the
semantic role `agent' to the syntactic 
function `subject' is revised. The agent now has to be realized as part of a
``von''--phrase, e.g. ``dieser Termin wird {\em von dem Besucher} gew\"unscht.''
(word--for--word: ``this date is whished {\em by the visitor} (dative
object)''). Furthermore, the 
object checks\linebreak whether the previously uttered part of the sentence includes
some of the revised material (i.e., whether the object itself has participated
in uttering). If this is the case, it sends an error message up to the
uppermost object of the hierarchy that actually is engaged in uttering. This
object is able to synchronize global repair. Up to now, we just realized a
simple repair strategy that consists of repeating the concerned parts of the
utterance, e.g. ``der Besucher \ldots \"ah \ldots dieser Termin wird {\em von dem
Besucher} gew\"unscht''.}

\vspace{-4mm}
\section*{\protect\large\bf 4 DISCUSSION}
\label{sec-4}
\vspace{-2mm}

This paper proposes a pro\-cess\-ing--con\-form\-ing default handler for
generation realizing repair by non--mon\-oton\-ic changes. We provide the system
with default descriptions. The set of default preconditions expresses possible
reasons for dead--end situations. A default is triggered, if the preconditions
match the current situation and the certainty value of the default exceeds the
predefined threshold. The default body is expressed in terms of the missing
input specification in order to make the system work homogeneously. We have
verified the advantages of pro\-cess\-ing--con\-form\-ing default handling by
implementing a default handler for VM--GEN.

As future work, we will extend the default preconditions towards
handling complex contextual information. We will apply default handling to
microplanning and lexical choice within VERBMOBIL. With respect to a sophisticated
output, we aim to combine VM--GEN with a flexible repair component.

The system VM--GEN is used in the VERBMOBIL scenario for multilingual
generation (English, German, and Japanese). We mean by multilinguality that 
the same processing is applied for different languages. In the
underlying knowledge sources lan\-guage--spe\-cif\-ic constraints are defined. 
Default handling can be easily adapted to the requirements of multilingual
generation by using lan\-guage--spe\-cif\-ic de\-fault--de\-scrip\-tions. 

For all knowledge sources the question arises how knowledge can be shared. We
intend to use {\em core knowledge sources} for representing common
phenomena. The core set of default descriptions for English and German,
e.g., contains the description of a reaction to a missing number value for a
noun. We aim to develop an efficient storing mechanism using a hierarchy of
locally intersecting core descriptions.

\vspace{-3mm}
\section*{\large\bf References}
\begin{small}
\vspace{-2mm}
 \mybib{Appelt, D. (1985). {\em Planning English Sentences}. Cambridge, MA:
Cambridge University Press.}

 \mybib{Block, H.--U., Bosch, P., Engelkamp, J., v.\ Hahn, W., Hauenschild,
C., H\"oge, H., Rohrer, C., Tillmann, H., G., Wahlster, W., Waibel, A. (1992). {\em
Wissenschaftliche Ziele und Netzpl\"ane f\"ur das VERBMOBIL--Projekt}.
Technical report, German Research Center for Artificial Intelligence (DFKI
GmbH), Saarbr\"ucken, Germany, 1992. }

 \mybib{Bock, J., and Warren, R. (1985). Conceptual accessibility and syntactic
structure in sentence formulation. {\em Cognition}, {\em 21}, 47-67.}

 \mybib{De~Smedt, K. (1990). {\em Incremental Sentence Generation: a Computer Model of
Grammatical Encoding}. PhD thesis, Nijmegen Institute for Cognition Research and
Information Technology, Nijmegen, NICI TR No 90--01.}

 \mybib{Finkler, W. (1994). {\em Performing Self--Corrections During
Incremental Natural Language Generation}. Document, German Research Center for
Artificial Intelligence (DFKI GmbH), Saarbr\"ucken, Germany, 1994. to appear.}

 \mybib{Joshi, A. (1985). {\em An Introduction to TAGs}. Technical Report
MS-CIS-86-64, LINC-LAB-31, Department of Computer and Information Science,
Moore School, University of Pennsylvania.}

 \mybib{Kameyama, M., Ochitani, R., and Peters, S. (1991). {\em Resolving
Translation Mismatches With Information Flow}. 29th Annual Meeting of the
Association for Computational Linguistics, Berkeley, CA, pp.\ 193-200.}

 \mybib{Kilger, A. and Finkler, W. (1994). {\em TAG--based Incremental
Generation}. Technical report, German Research Center for Artificial
Intelligence (DFKI GmbH), Saarbr\"ucken, Germany. to appear.}

 \mybib{Kilger, A. (1994). Using UTAGs for Incremental and Parallel
Generation. {\em Computational Intelligence}. to appear.}

 \mybib{Kitano, H. (1990). {\em Incremental Sentence Production with a Parallel
Marker-Passing Algorithm}. 13th International Conference on Computational
Linguistics, Helsinki, Finland, pp.\ 217 - 222.}

 \mybib{Meteer, M. (1990). {\em The ''Generation Gap'': the Problem of
Expressibility in Text Planning}. Department of Computer and Information
Science, University of Massachusetts, Amherst, MA, BBN Report No.\ 7347.}

 \mybib{Ward, N. (1991). {\em A Flexible, Parallel Model of Natural Language
Generation}. PhD thesis, Computer Science Division (EECS), University of
California, Berkeley, CA, Report No. UCB/CSD 91/629.}

\end{small}

\end{document}